\documentclass[sigconf]{acmart}
\usepackage{popets}
\usepackage{multirow}
\usepackage{threeparttable}
\setcopyright{popets}
\copyrightyear{YYYY}
\acmYear{YYYY}
\acmVolume{YYYY}
\acmNumber{X}
\acmDOI{XXXXXXX.XXXXXXX}
\acmISBN{}
\acmConference{Proceedings on Privacy Enhancing Technologies}
\usepackage{tabularx}
\usepackage[most]{tcolorbox}
\usepackage{xcolor} % For color definitions
\usepackage{graphicx}
\usepackage{subcaption}
\usepackage{hyperref}
\usepackage{multirow}
\usepackage{amsmath,amsfonts}
\usepackage{xspace}
\usepackage{tcolorbox}

\usepackage{amsthm}
\usepackage{subcaption}  
\usepackage{algorithm}
\usepackage{algpseudocode}
\usepackage{amsmath}
\usepackage{float}
\usepackage{booktabs}
\usepackage{titlesec}
\usepackage[utf8]{inputenc} % allow utf-8 input
\usepackage[T1]{fontenc}    % use 8-bit T1 fonts
\usepackage{hyperref}       % hyperlinks
\usepackage{url}            % simple URL typesetting
\usepackage{booktabs}       % professional-quality tables
\usepackage{amsfonts}       % blackboard math symbols
\usepackage{nicefrac}       % compact symbols for 1/2, etc.
\usepackage{microtype}      % microtypography
\usepackage{xcolor}   
\usepackage{enumitem}
\usepackage[title]{appendix}
\begin{document}
\title{Are Neuromorphic Architectures Inherently Privacy-preserving?\\ An Exploratory Study}
\author{Ayana Moshruba}
\affiliation{%
  \institution{George Mason University}
  \city{}
  \state{}
  \country{}}
\email{amoshrub@gmu.edu}
\author{Ihsen Alouani}
\affiliation{%
  \institution{Centre for Secure Information Technologies (CSIT)\\Queen's University Belfast}
  \city{}
  \country{}}
\email{i.alouani@qub.ac.uk}
\author{Maryam Parsa}
\affiliation{%
  \institution{George Mason University}
  \city{}
  \state{}
  \country{}}
\email{mparsa@gmu.edu}
\renewcommand{\shortauthors}{Moshruba et al.}
\begin{abstract}
\noindent
While machine learning (ML) models are becoming mainstream, including in critical application domains, concerns have been raised about the increasing risk of sensitive data leakage. Various privacy attacks, such as membership inference attacks (MIAs), have been developed to extract data from trained ML models, posing significant risks to data confidentiality. %Membership Inference Attack (MIA) is one of the foundational privacy attacks. %MIA aims to determine weather  a data sample has been used to train a victim model.
%While the privacy issues have been extensively investigated in the context of traditional Artificial Neural Networks (ANNs), it remains largely unexplored in neuromorphic architectures, such as Spiking Neural Networks (SNNs).
While the predominant work in the ML community considers traditional Artificial Neural Networks (ANNs) as the default neural model, neuromorphic architectures, such as Spiking Neural Networks (SNNs) have recently emerged as an attractive alternative mainly due to their significantly low power consumption. These architectures process information through discrete events, i.e., spikes, to mimic the functioning of biological neurons in the brain.  %However, little work has been dedicated to investigate their privacy-preserving properties. 
While the privacy issues have been extensively investigated in the context of traditional ANNs, they remain largely unexplored in neuromorphic architectures, and little work has been dedicated to investigate their privacy preserving properties. %; it is still not clear if SNNs' unique computing paradigm has any privacy-preserving advantage. 
%SNNs process data in an event-driven manner, activating neurons only when incoming stimuli exceed specific thresholds, which enhances temporal data handling, reduces energy consumption, and accelerates response times.
 %In this paper, we investigate the privacy-preserving properties of SNNs compared to ANNs by performing a series of MIAs  across multiple datasets such as MNIST, F-MNIST, Iris, Breast Cancer, CIFAR-10, CIFAR-100 and ImageNet. Moreover, We explore the effects of different learning algorithms such as: surrogate gradient and evolutionary learning; and different programming frameworks (snnTorch, TENNLab, and LAVA), and various parameters within these algorithms on the resilience of SNNs against MIA. Our comprehensive experiments across diverse datasets reveal that SNNs consistently demonstrate superior privacy preservation compared to ANNs, with evolutionary algorithms further enhancing their resilience.
In this paper, we investigate the question whether SNNs have inherent privacy preserving advantage.
Specifically, we investigate SNNs' privacy properties through the lens of MIAs across diverse datasets, in comparison with ANNs. 
We explore the impact of different learning algorithms (surrogate gradient and evolutionary learning), programming frameworks (snnTorch, TENNLab, and LAVA), and various parameters on the resilience of SNNs against MIA. Our experiments reveal that SNNs demonstrate consistently superior privacy preservation compared to ANNs, with evolutionary algorithms further enhancing their resilience. For example, on the CIFAR-10 dataset, SNNs achieve an AUC as low as 0.59 compared to 0.82 for ANNs, and on CIFAR-100, SNNs maintain a low AUC of 0.58, whereas ANNs reach 0.88.
% The Attack Area Under the Receiver Operating Characteristic Curve (AUC-ROC) for SNN is 16.39\% lower on MNIST, 17.14\% lower on CIFAR-10, and 11.48\% lower on ImageNet compared to ANN. %This inherent privacy protection of SNN models can be attributed to the non-differentiability or discontinuity in them, which results in a lower correlation between the model and individual data points, thereby restricting data leakage. 
%In addition, we also observed that the networks trained with evolutionary learning algorithms show higher resiliency towards MIA than the gradient based learning algorithm. 
Furthermore, we investigate the privacy-utility trade off through Differentially Private Stochastic Gradient Descent (DPSGD) observing that SNNs incur a notably lower accuracy drop than ANNs under equivalent privacy constraints. %, a training method that introduces calibrated noise to the gradients. %This approach provides mathematically quantifiable privacy guarantees to both SNNs and ANNs.
% Our experiments demonstrate that SNNs exhibit a lower drop in accuracy compared to ANNs for the same level of privacy guarantees. For example, with privacy budget range of 0.17-1.03, the average accuracy drop for SNN is 7.78\% on Fashion-MNIST, significantly lower than the 23.71\% drop observed in ANN. These results align with our findings from the MIA experiments. Our research highlights significantly higher privacy-preserving potential of SNNs, providing an incentive for their adoption in sensitive applications. \textcolor{blue}{\textit{Ihsen}-- please illustrate with some numbers--}
% \textcolor{orange}{\textit{Ayana}-- done--}
\end{abstract}

\maketitle
\section{Introduction}\label{sec:intro}
As ML systems become more sophisticated and widespread, individuals are increasingly relying on these systems, entrusting them with personal and professional data. Consequently, the risk of sensitive information exposure~\cite{salomon2012data} is growing significantly in multiple sectors \cite{bertino2016data} including healthcare~\cite{abouelmehdi2017big}, finance~\cite{tripathi2020financial}, national security~\cite{thuraisingham2002data}, education~\cite{florea2020big} and consumer services~\cite{lee2015personalized}. It is particularly alarming in fields such as healthcare, where the confidentiality of patient data is extremely sensitive, as a breach could result in severe personal and financial implications, which can affect patient care and institutional credibility~\cite{luo2018privacyprotector, jm2018data, hipaa}. In finance, the integrity of financial transactions and records is fundamental for maintaining market stability and preventing fraud~\cite{yu2021corporate}, while in national security, safeguarding classified information is essential to protect national interests and prevent threats to public safety~\cite{kim2010security}.

This has led to the development of various privacy attacks targeting ML models to extract sensitive information, including Model Inversion Attacks~\cite{fredrikson2015model}, Attribute Inference Attacks~\cite{gong2018attribute}, Model Stealing Attacks~\cite{juuti2019prada}, and Membership Inference Attacks (MIAs)~\cite{shejwalkar2021membership}. In MIAs, an adversary seeks to ascertain if a specific data point was part of the dataset used to train the model. This intrusion risks exposing classified information about individuals in the training dataset, potentially compromising personal data confidentiality~\cite{shokri}.

Designed to replicate the dynamic behavior of biological neurons~\cite{ghosh2009spiking}, SNNs process information through discrete, temporally encoded spikes~\cite{roy2019towards}, enabling them to handle time-sensitive data efficiently~\cite{hong2017tafc}. Their suitability for edge computing~\cite{shi2016edge} and resource-constrained environments further enhances their value, as SNNs effectively process real-world spatiotemporal patterns~\cite{wang2018learning}. This capability positions SNNs as a promising alternative to traditional neural networks for applications requiring dynamic, real-time data processing. While the security of SNNs has been investigated in the literature~\cite{date21,ijcnn20}, relatively little attention has been given to their privacy-preserving capabilities.

This work addresses the privacy concerns associated with SNNs through a structured investigation of three core areas: (i) the resilience of ANN and SNN models to Membership Inference Attacks (MIAs), (ii) the factors influencing the privacy-preserving properties of SNNs, and (iii) the privacy-utility trade-off in ANN and SNN models using the DPSGD algorithm.

We consider that the potential resilience of SNNs against MIAs is based on two key aspects. Firstly, the non-differentiable and discontinuous nature of SNNs may weaken the correlation between the model and individual data points, making it more challenging for an attacker to identify the membership of a particular data point in the training set~\cite{meng2022training}. Secondly, the unique encoding mechanisms employed by SNNs introduce an additional layer of stochasticity~\cite{olin2021stochasticity} and variability to the data representation. This added complexity can make it more difficult for an attacker to deduce the unique characteristics of individual data points, thereby making them more indistinguishable.

Investigating the resilience of SNNs against MIAs, our experimental results consistently demonstrate that SNNs exhibit higher resilience to MIAs across the datasets including MNIST, F-MNIST, Iris, Breast Cancer, CIFAR-10, and CIFAR-100. This is evidenced by the lower Area Under the Curve(AUC) values for the Receiver Operating Characteristic(ROC) curves in SNNs compared to their ANN counterparts. Furthermore, our exploration domain encompasses various learning algorithms (surrogate gradient-based and evolutionary learning), programming frameworks (snnTorch, TENNLab, and LAVA), and a wide range of parameters within them, providing a comprehensive analysis of the factors influencing the inherent privacy preserving properties of SNNs. This in-depth exploration indicated that evolutionary learning algorithms shown to boost this resilience more effectively compared to the gradient based methods.

In order to enhance data privacy and explore the compromises between privacy and utility, we study the implementation of the DPSGD algorithm as a privacy defense mechanism~\cite{xu2021dp}. This introduces controlled noise into the training process, making it harder for attackers to infer the presence of specific data points. However, improved privacy often comes at the cost of reduced model performance, known as the privacy-utility trade-off ~\cite{song2013stochastic}. Through the experiments, we observe that SNNs exhibit a notably lower performance drop compared to ANNs for the same level of privacy guarantee. This finding further reinforces our hypothesis regarding the inherent privacy-preserving properties of SNNs. 

This paper offers the following notable contributions in the field of data privacy, particularly in the context of SNNs:

\begin{itemize}[leftmargin=4mm]
\item SNNs exhibit higher resilience against MIAs compared to ANNs, with lower AUC scores on CIFAR-10 (SNN: 0.59 vs. ANN: 0.82) and CIFAR-100 (SNN: 0.58 vs. ANN: 0.88), highlighting their potential as a more secure alternative in privacy sensitive applications.
\item Evolutionary learning algorithms outperform gradient based methods in MIA resilience, maintaining a consistent AUC of ~0.50 across all parameters for Iris and Breast Cancer datasets, compared to 0.57 and 0.55 AUC scores for gradient-based algorithms, respectively.
\item Privacy-utility trade off analysis revealing that SNNs incur a lower accuracy drop compared to ANNs when applying DPSGD: For F-MNIST, with privacy guarantees ranging from 0.22 to 2.00, the average accuracy drop is 12.87\% for SNNs comparatively lower than the 19.55\% drop observed in ANNs.
\end{itemize}

We emphasize that while this investigation highlights SNNs' enhanced privacy characteristics, our findings specifically address privacy preservation applications. The architectural properties of SNNs that enable efficient hardware implementation and reduce computational overhead make them particularly appealing for resource constrained environments. However, these findings do not suggest an overall superiority of SNNs over ANNs across different application domains. Rather, we base this work on the intuition that SNNs' unique information processing mechanisms may offer specific advantages in privacy preservation, which warrants systematic investigation in this particular domain. The implementation of all experiments conducted in this study is publicly available at \href{https://github.com/AyanaMoshruba/Neuromorphic_Privacy.git}{\textcolor{blue}{this repository}}.

\section{Background}
\begin{figure*}[ht]
\centering
\includegraphics[width=1\linewidth]{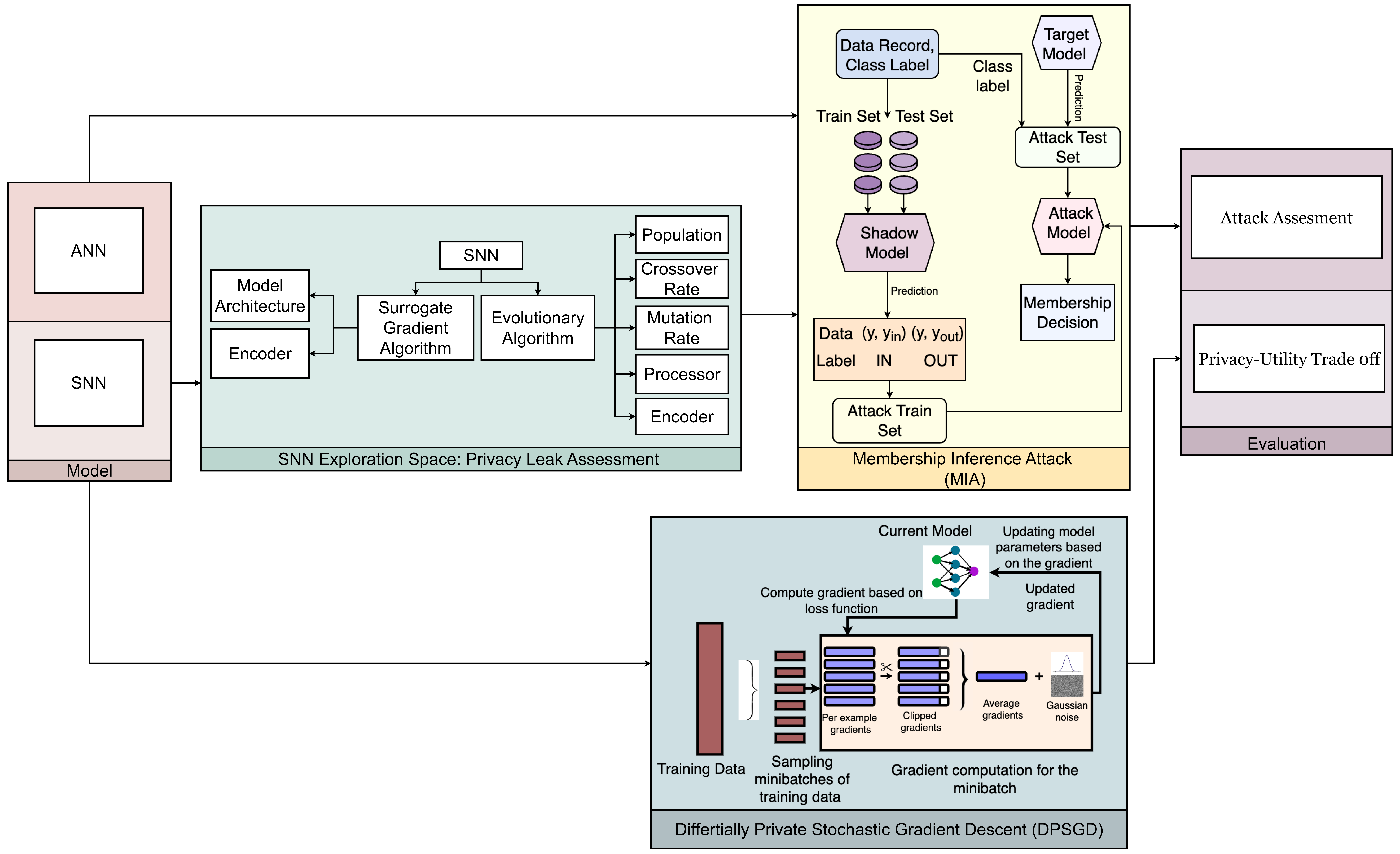}
\caption{Architectural overview of the experimental framework. The methodology has 3 components: (i) comparative privacy assessment between ANNs and SNNs through MIA (yellow), (ii) exploration of SNN specific privacy characteristics considering surrogate gradient and evolutionary algorithms (green), and (iii) Comparative evaluation of privacy-utility trade offs through DPSGD (blue)} 
\Description{} 
\label{fig:flow}
\end{figure*}
\subsection{Neuromorphic Architecture}
Neuromorphic architectures ~\cite{markovic2020physics} are designed to mimic the neural structures and functionalities of the biological brain, offering an alternative to the traditional von Neumann computing systems ~\cite{aspray1990john}. These architectures integrate memory and processing units, enabling massive parallel processing ~\cite{nassi2009parallel} and event driven computation ~\cite{hwang2000predictive}. By modeling neurons and synapses that communicate via discrete spikes or events ~\cite{liu2021low}, neuromorphic systems can process data asynchronously, reducing power consumption and latency. Hardware implementations like IBM's TrueNorth ~\cite{akopyan2015truenorth} and Intel's Loihi ~\cite{davies2018loihi} demonstrate the potential for scalable, energy efficient architectures capable of performing complex tasks with a structure resembling biological neural networks.

SNNs constitute the foundational architecture of neuromorphic systems, operating through a mechanism where neurons accumulate membrane potential over time and generate discrete spike events upon reaching a threshold potential. This temporal encoding paradigm fundamentally differentiates SNNs from traditional neural networks, as information is encoded in both spike timing and frequency. The computational models of SNNs span from elementary integrate-and-fire neurons ~\cite{auge2021survey} to sophisticated biologically inspired implementations such as the Hodgkin Huxley model, which incorporates detailed ionic conductance dynamics. The event driven nature of spike-based computation intrinsically aligned with neuromorphic hardware architectures, enabling efficient asynchronous processing. This integration of temporal dynamics and synaptic plasticity in SNNs facilitates adaptive learning mechanisms particularly suited for applications demanding real time processing and minimal latency ~\cite{schuman2022opportunities}.

The architectural distinctiveness of SNNs, characterized by spike based information encoding and asynchronous processing, introduces computational complexities that potentially influence their susceptibility to privacy attacks. While traditional ANNs demonstrate vulnerability to Membership Inference Attacks through their deterministic output patterns and continuous gradients, SNNs operate via discrete, non-differentiable spike events ~\cite{zhang2018plasticity}. The temporal dynamics of spike generation and the stochastic nature of neuronal activation contribute to output variability. The membrane potential accumulation in SNN neurons, coupled with probabilistic firing thresholds, generates diverse spike patterns for comparable inputs. This intrinsic variability and the discontinuous activation characteristics inherent to SNNs potentially obscure input-output relationships, thereby complicating the pattern recognition essential for successful MIAs. Although these properties, particularly when combined with differential privacy techniques like DPSGD, suggest enhanced privacy preservation capabilities, such assumptions require rigorous validation. This investigation examines whether the spike based computational paradigm of SNNs exhibits superior resilience to performance degradation under privacy constraints compared to traditional ANNs.
\subsection{Membership Inference Attack (MIA)}
constitute a class of privacy vulnerabilities that enable adversaries to determine whether specific data points were used in a model's training dataset. These attacks exploit differential behavioral patterns in model responses between training and non-training samples ~\cite{rahman2018membership}. Neural networks typically demonstrate heightened prediction confidence and distinctive error distributions for previously encountered training samples compared to unseen data ~\cite{nguyen2015deep}. Through systematic analysis of these response characteristics, adversaries can extract sensitive training set information, potentially compromising data privacy ~\cite{de2020overview} in domains handling confidential data.
The implementation of MIAs encompasses both the development of specialized attack models and the application of statistical inference methods to differentiate between model responses to training and non-training samples. The efficacy of these attacks correlates strongly with the degree of model overfitting and the distinctiveness of individual sample responses. Beyond immediate privacy implications, MIAs serve as indicators of model generalization deficiencies ~\cite{gomm2000case}, highlighting vulnerabilities in the neural network architecture.
\subsection{Differentially Private Stochastic Gradient Descent (DPSGD)}
Differential Privacy (DP) ~\cite{dwork2006differential} establishes a mathematical framework that quantifies and bounds the privacy risk when operating on sensitive data. The framework ensures that statistical queries on a dataset remain nearly unchanged regardless of the inclusion or exclusion of any individual record, providing formal privacy guarantees. This privacy guarantee is parameterized by \(\epsilon\) (privacy budget) and \(\delta\) (failure probability), formalized through the following inequality:
 \[ P(M(D) \in S) \leq e^\epsilon P(M(D') \in S) + \delta \]
where \(D\) and \(D'\) are datasets differing by one element, \(M\) is a randomized algorithm, and \(S\) is a subset of possible outputs. This inequality limits the probability of \(M\) producing an output within \(S\) when applied to \(D\) is limited by the exponential of \(\epsilon\). This probability is then multiplied by the probability of \(M\) producing the same output from \(D'\). Additionally, it is adjusted by a small term \(\delta\).
In this context, \(\epsilon\)  controls the sensitivity of the output to variations in the input, where smaller values of \(\epsilon\) indicate enhanced privacy by limiting the permissible changes in output probabilities. \(\delta\), ideally a small value near zero, accounts for the rare cases where this privacy guarantee might fail, thus ensuring that changes to any single data point in \(D\) minimally affect the output, reinforcing data privacy across the dataset.

DPSGD implements these theoretical guarantees in the context of neural network training by incorporating calibrated noise into the gradient computation process. In DPSGD, the key step involves perturbing the gradients computed during each training iteration with noise that is calibrated to the sensitivity of the function being optimized. This sensitivity measures how much the output of a function can change in response to changes in its input, which in the context of machine learning, translates to how much a single training example can influence the overall model parameters.
The noise added is typically drawn from a Gaussian distribution ~\cite{kassam2012signal}, scaled according to \( \epsilon \) and the desired level of privacy guarantee, \( \delta \). 
The function of DPSGD can be expressed mathematically as:
\[
\theta_{t+1} = \theta_t - \eta \left(g_t + \mathcal{N}(0, \sigma^2 \mathbf{I})\right)
\]
where \(\theta_t\) represents the model parameters at iteration \(t\), \(\eta\) is the learning rate, \(g_t\) is the gradient of the loss function with respect to \(\theta_t\), clipped to a norm bound \(C\), and \(\mathcal{N}(0, \sigma^2 \mathbf{I})\) denotes the Gaussian noise added to the gradient, with \(\sigma\) being determined by \(C\), \(\epsilon\), and \(\delta\).

\section{Methodology}

This investigation examines the comparative privacy resilience of SNNs and ANNs through a systematic experimental framework comprising three distinct phases:
(1) assessment of privacy vulnerabilities through MIA in both architectures,
(2) analysis of SNN specific privacy characteristics across diverse algorithmic implementations, and
(3) evaluation of privacy-utility trade offs through the implementation of DPSGD.
The experimental methodology and interrelationships between architectural components are illustrated in Figure~\ref{fig:flow}.
\subsection{Comparison of Privacy Vulnerability between ANNs and SNNs}
The comparative privacy risk assessment utilizes MIAs across equivalent ANN and SNN architectures, implementing convolutional baseline models, ResNet18, and VGG16 configurations. The experimental framework employs shadow models trained on labeled datasets (Figure~\ref{fig:flow}, top-right) to emulate the target model characteristics, while attack models are developed to ascertain training set membership of individual data points. Privacy vulnerability is quantified through AUC metrics for both architectures, enabling systematic comparison of their susceptibility to MIAs (Figure~\ref{fig:flow}, yellow block).
\subsection{Algorithmic Exploration within the SNN Architecture}
The second phase examines SNN privacy resilience across diverse algorithmic implementations through three distinct frameworks: Surrogate Gradient Algorithm, Evolutionary Algorithm, and Intel's LAVA framework.
The Surrogate Gradient implementation evaluates the baseline SNN model using three distinct encoding mechanisms from the snnTorch~\cite{snntorch} library: Delta, Latency, and Delta Modulation (Figure~\ref{fig:flow}, left section, "SNN Exploration Space"). This analysis examines how encoding methods affect privacy vulnerability by correlating spike generation mechanisms with MIA efficacy.

The Evolutionary Algorithm implementation, utilizing the TennLab framework, modulates architectural parameters including population size, crossover rate, mutation rate, processor configurations, and encoder settings. This parameter space exploration evaluates how architectural variations influence SNN susceptibility to MIAs (Figure~\ref{fig:flow}, green block).
\subsection{Privacy-Utility Trade-off Analysis}
The final experimental phase implements DPSGD across both ANN and SNN architectures ((Figure~\ref{fig:flow}, bottom) to quantify the privacy-utility trade-offs. The DPSGD implementation incorporates Gaussian noise during gradient computation, establishing differential privacy guarantees while measuring the corresponding impact on model performance. The analysis spans convolutional, ResNet18, and VGG16 architectures, evaluating both model accuracy and resilience through attack AUC metrics under equivalent privacy constraints (Figure~\ref{fig:flow}, blue block).
\section{Experimental Framework and Setup}
\subsection{Dataset and Model Architecture}
\label{subsec:arch}
\begin{table}[t]
\small
\setlength{\tabcolsep}{4pt}  % Reduce column spacing (default is 6pt)
\begin{center}
\caption{Model Architectures and Configurations}
\label{tab:architectures}
\begin{tabular}{p{1.4cm}p{0.8cm}p{3.8cm}p{1.2cm}}  % Reduced width of Structure column
\hline
Network & Variant & Structure & Parameters* \\
\hline
Baseline & ANN & $\bullet$ 2 Conv (32,64 filters) & \multirow{2}{*}{$\sim$2.3M} \\
(ConvNet) & & $\bullet$ 2 MaxPool & \\
 & & $\bullet$ 2 FC (1000, num\_classes) & \\
 & & $\bullet$ ReLU & \\
\cline{2-4}
 & SNN & $\bullet$ 2 Conv (32,64 filters) & \multirow{2}{*}{$\sim$2.3M} \\
 & & $\bullet$ 2 MaxPool & \\
 & & $\bullet$ LIF neurons & \\
 & & $\bullet$ Temporal processing & \\
\hline
ResNet18 & ANN & $\bullet$ 4 BasicBlocks (64$\to$512) & \multirow{2}{*}{$\sim$11.7M} \\
 & & $\bullet$ GroupNorm & \\
 & & $\bullet$ Skip connections & \\
 & & $\bullet$ Adaptive pool & \\
\cline{2-4}
 & SNN & $\bullet$ 4 BasicBlocks (64$\to$512) & \multirow{2}{*}{$\sim$11.7M} \\
 & & $\bullet$ GroupNorm+BNTT & \\
 & & $\bullet$ Spike residuals & \\
 & & $\bullet$ Membrane threshold & \\
\hline
VGG16 & ANN & $\bullet$ 13 Conv (64$\to$512) & \multirow{2}{*}{$\sim$138M} \\
 & & $\bullet$ 5 MaxPool & \\
 & & $\bullet$ 3 FC (4096, classes) & \\
 & & $\bullet$ GroupNorm, ReLU & \\
\cline{2-4}
 & SNN & $\bullet$ 13 Conv (64$\to$512) & \multirow{2}{*}{$\sim$138M} \\
 & & $\bullet$ 5 AvgPool & \\
 & & $\bullet$ Binary spikes & \\
 & & $\bullet$ Membrane reset & \\
\hline
\end{tabular}
{\small
\begin{flushleft}
*Params vary with input channels (1/3) and classes (10/100)\\
\end{flushleft}}
\end{center}
\end{table}
The proposed method is evaluated on both image and tabular datasets. For image classification tasks, MNIST~\cite{lecun2010mnist} and Fashion-MNIST~\cite{xiao2017fashion}, comprising 28×28 grayscale images across 10 classes, are utilized. CIFAR-10~\cite{recht2018cifar} and CIFAR-100~\cite{sharma2018analysis}, containing 32×32 RGB images with 10 and 100 classes, respectively, are also included. Two tabular datasets, Iris~\cite{omelina2021survey}, consisting of 4 features and 3 classes, and Breast Cancer~\cite{misc_breast_cancer_14}, comprising 30 features and 2 classes, are used.

The experimental evaluation encompasses both image and tabular datasets. The image classification tasks utilize MNIST~\cite{lecun2010mnist} and Fashion-MNIST~\cite{xiao2017fashion}, comprising 28×28 grayscale images across 10 classes, and CIFAR-10~\cite{recht2018cifar} and CIFAR-100~\cite{sharma2018analysis}, containing 32×32 RGB images with 10 and 100 classes, respectively. The tabular datasets include Iris~\cite{omelina2021survey} (4 features, 3 classes) and Breast Cancer~\cite{misc_breast_cancer_14} (30 features, 2 classes).

The architectural implementations, as detailed in Table~\ref{tab:architectures}, are composed of three model configurations that are adapted for both ANN and SNN frameworks. The baseline architecture is implemented with dual convolutional layers for image processing and fully connected layers for tabular data analysis. ResNet18 is constructed with four basic blocks that incorporate group normalization and residual connections, while VGG16 is designed with 13 convolutional layers, progressively expanding from 64 to 512 channels, and culminating in three fully connected layers.

In the SNN variants, ReLU activations are replaced with leaky integrate-and-fire (LIF) neurons, and temporal processing mechanisms are incorporated. The implementation framework employs PyTorch~\cite{ketkar2021introduction} for ANN architectures, while snnTorch is used for the baseline model and SpikingJelly~\cite{spkjly} is utilized for ResNet and VGG16 configurations.
\subsection{MIA}
\begin{figure}[ht!]
    \centering    \includegraphics[width=0.8\linewidth, height=\linewidth]{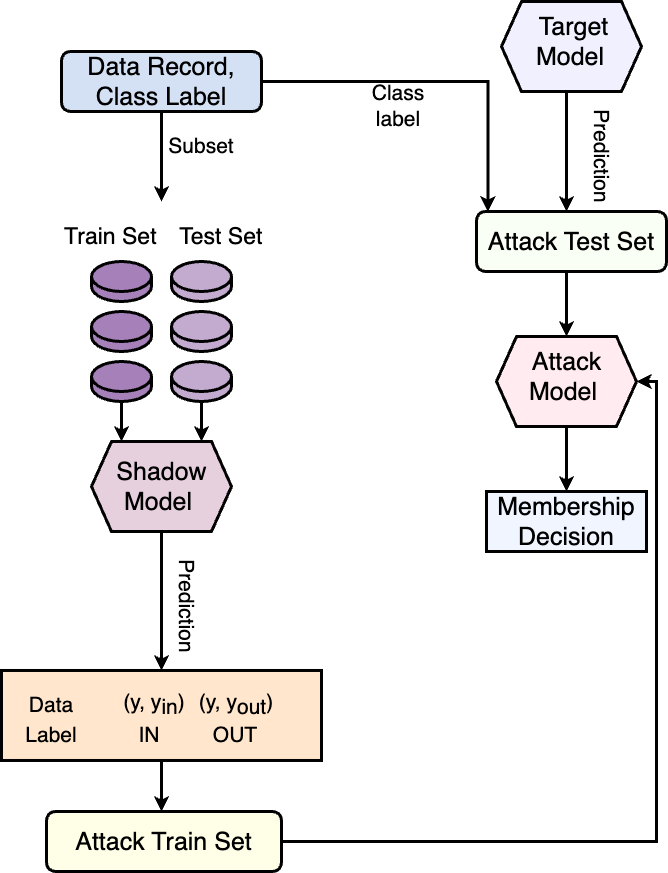}
    \caption{Membership Inference Attack(MIA) Framework}
    \Description{}    
    \label{fig:mia_24}
\end{figure}
The MIA framework employs a dual model architecture as illustrated in Figure~\ref{fig:mia_24}. The experimental setup includes a 'target' model and a 'shadow' model, where we evaluate three different architectures: baseline model, ResNet18, and VGG16. The shadow model simulates the target model's functionality by training on 80\% of the corresponding dataset while maintaining architectural parity. For the attack phase, we employ a Support Vector Machine (SVM) as our attack model, chosen for its effectiveness as a binary classifier to determine whether a data point is a member or non-member of the training set. We evaluate the models by querying with their respective training and test sets, labeling the responses as 'IN' if the data was in the model's training set and 'OUT' otherwise. To reduce classifier bias towards more frequent classes, we implement undersampling of the predominant class. The SVM attack model, trained on these labeled predictions from the shadow model, provides a quantified measure of the model's susceptibility to MIA through its ability to accurately classify membership status.
\subsection{\textbf{DPSGD}}
\begin{figure*}[ht!]
    \centering    \includegraphics[width=0.6\linewidth]{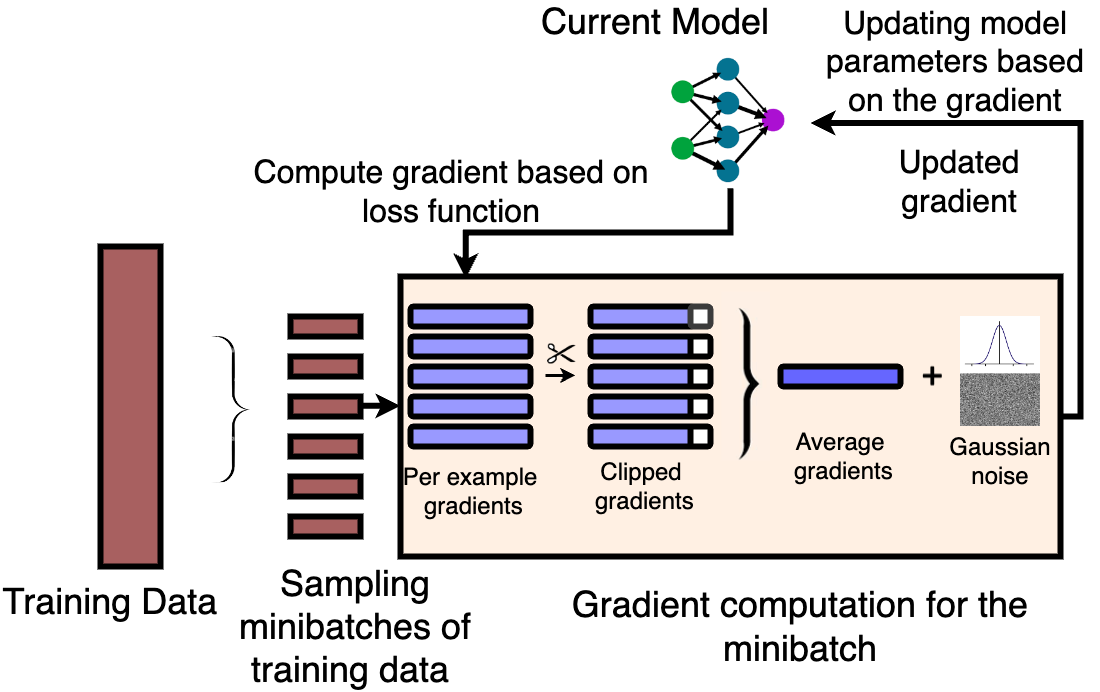}
    \caption{Differential Privacy via Stochastic Gradient Descent (DPSGD) Algorithm}
    \Description{}
    \label{fig:dpsgd}  
\end{figure*}
The DPSGD implementation utilizes the Opacus Privacy Engine \cite{yousefpour2021opacus}, as illustrated in Figure~\ref{fig:dpsgd}. The DPSGD algorithm ~\ref{alg:dpsgd} begins by selecting a minibatch $L_t$ uniformly sampled from the set of indices $\{1, \dots, N\}$ with size $L$. For each data point $x_i$ in the minibatch, the gradient of the loss function $\mathcal{L}(\theta_t, x_i)$ with respect to the parameters $\theta_t$ is computed as $g_t(x_i)$. These gradients are then clipped using an $L2$ norm operation, $\hat{g}_t(x_i) = g_t(x_i) / \max(1, \|g_t(x_i)\|_2 / C)$. Gaussian noise $\mathcal{N}(0, \sigma^2 C^2 I)$ is added to the average of the clipped gradients, with $\sigma$ being the noise scale, to ensure differential privacy. The noisy gradient $\tilde{g}_t$ is used to update the model parameters: $\theta_{t+1} = \theta_t - 
 \eta_t \tilde{g}_t$.

To investigate the privacy-utility tradeoff, we define the same target privacy budget ($\varepsilon$) of range 0.1-2.00, while fixing the privacy compromise $\delta=1e-5$ and evaluate the utility of the SNN architectures comparatively with their ANN counterparts.

\begin{algorithm}[H]
\caption{DPSGD Algorithm}
\label{alg:dpsgd}
\begin{algorithmic}[1]
\State \textbf{Input:} Dataset $\{x_1, \dots, x_N\}$, loss function $\mathcal{L}(\theta) = \frac{1}{N}\sum_{i=1}^N \mathcal{L}(\theta, x_i)$
\State \textbf{Parameters:} Learning rate $\eta$, noise scale $\sigma$, batch size $L$, gradient norm bound $C$
\State Initialize $\theta_0$ randomly
\For{$t = 1$ \textbf{to} $T$}
\State Take a random sample $L_t$ with sampling probability $\frac{L}{N}$
\For{each $i \in L_t$}
\State Compute gradient $g_t(x_i) \leftarrow \nabla_\theta \mathcal{L}(\theta_t, x_i)$
\State Clip gradient $g_t(x_i) \leftarrow \frac{g_t(x_i)}{\max(1, \frac{\|g_t(x_i)\|_2}{C})}$
\EndFor
\State Add noise $\tilde{g}_t \leftarrow \frac{1}{L} \left(\sum_{i \in L_t} g_t(x_i) + \mathcal{N}(0, \sigma^2 C^2 I)\right)$
\State Update parameters $\theta_{t+1} \leftarrow \theta_t - \eta \tilde{g}_t$
\EndFor
\State \textbf{Output:} Return $\theta_T$ and compute privacy cost $(\epsilon, \delta)$
\end{algorithmic}
\end{algorithm}

% \begin{table}[htp]
% \centering
% \caption{DP-SGD Parameters}
% \label{tab:dpsgd_params}
% \begin{tabular}{lc}
% \toprule
% \textbf{Parameter} & \textbf{Value} \\
% \midrule
% Learning rate ($\eta$) & $0.1$ \\
% Noise multiplier ($\sigma$) & $1$ \\
% Privacy parameter ($\delta$) & $1e-5$ \\
% Gradient norm ($C$) & $1.0$ \\
% \bottomrule
% \end{tabular}
% \end{table}

\subsection{SNN Exploration Space}
\label{subsec:exploration}

In the SNN exploration space as shown in Figure~\ref{fig:flow}, we investigate two major learning algorithms: surrogate gradient based learning ~\cite{neftci2019surrogate} and evolutionary learning~\cite{schuman2020evolutionary}.

\subsubsection{\textbf{Surrogate Gradient based Learning:}}

Surrogate gradient methodologies facilitate SNN training through differentiable approximations of the non differentiable spiking activation function, enabling gradient based optimization techniques. These algorithms substitute the discrete spike function gradient with continuous, differentiable surrogates, permitting backpropagation based training while preserving the essential characteristics of spike-based computation. The experimental implementation employs two distinct approaches: the Arc tangent surrogate gradient algorithm ~\cite{1628884} implemented via snnTorch ~\cite{snn} framework, and the biologically-inspired Spike Timing Dependent Plasticity (STDP) ~\cite{caporale2008spike} learning mechanism through Intel's LAVA framework ~\cite{lava2023}.

\noindent{\textbf{SnnTorch Framework:}}

The snnTorch framework, built upon PyTorch, provides specialized implementations for Spiking Neural Networks. Our experimental framework implements the arc tangent learning algorithm, which provides differentiable approximations of the spiking activation function. This implementation is evaluated across MNIST, F-MNIST, Iris, and Breast Cancer datasets to assess architectural resilience against membership inference attacks.
In SNN architectures, the encoding mechanism determines the transformation of input signals into temporal spike patterns, fundamentally influencing the network's computational characteristics and performance metrics. Our investigation examines the relationship between encoding methodologies and privacy preservation characteristics across three distinct encoding schemes ~\cite{snntorch}:

\begin{itemize}
    \item \textit{Rate Encoding:} Transforms input features into spikes by representing each feature as a probability of spike occurrence at each time step, with the neuron's firing rate directly tied to the intensity of the input signal. For our experiments, we set number of steps to 10.
    \item \textit{Latency Encoding:} Encodes information based on the timing of spikes, where inputs with higher values result in earlier spikes, effectively using the temporal dimension to convey input magnitude. We used time step of 10 and set the RC constant,  \textit{Tau} to 5, and Threshold to 0.1 for our experiments.
    \item \textit{Delta Encoding:} Event-driven and produces spikes in response to changes in input features over time, making it adept at capturing dynamic variations in data. In our experiments, we set the threshold to 0.1.
\end{itemize}

\noindent{\textbf{LAVA Framework:}}

LAVA is an innovative open source software framework designed for neuromorphic computing ~\cite{roy2019towards}. It provides a versatile and flexible environment for developing, simulating, and deploying neuromorphic applications. By abstracting hardware details through a process-based model, LAVA enables the creation of scalable and modular systems that can operate asynchronously across various neuromorphic and conventional hardware platforms. The framework supports rapid prototyping and detailed optimization, making it accessible to both researchers and developers interested in leveraging the unique capabilities of neuromorphic technologies. LAVA aligns particularly well with Intel's Loihi chip ~\cite{davies2018loihi}, a specialized neuromorphic processor. LAVA provides a seamless interface for developing applications that can efficiently run on Loihi. The LAVA framework also supports Spike-Timing-Dependent Plasticity (STDP), a biological learning rule that adjusts synaptic strengths based on the precise timing of spikes. This mechanism allows SNNs to learn temporal patterns and adapt to dynamic inputs. In our experiment, we used framework to evaluate the model's resilience against privacy attack (MIA).

\begin{figure*}[ht]
    \centering
    \includegraphics[width=\linewidth, height=0.6\linewidth]{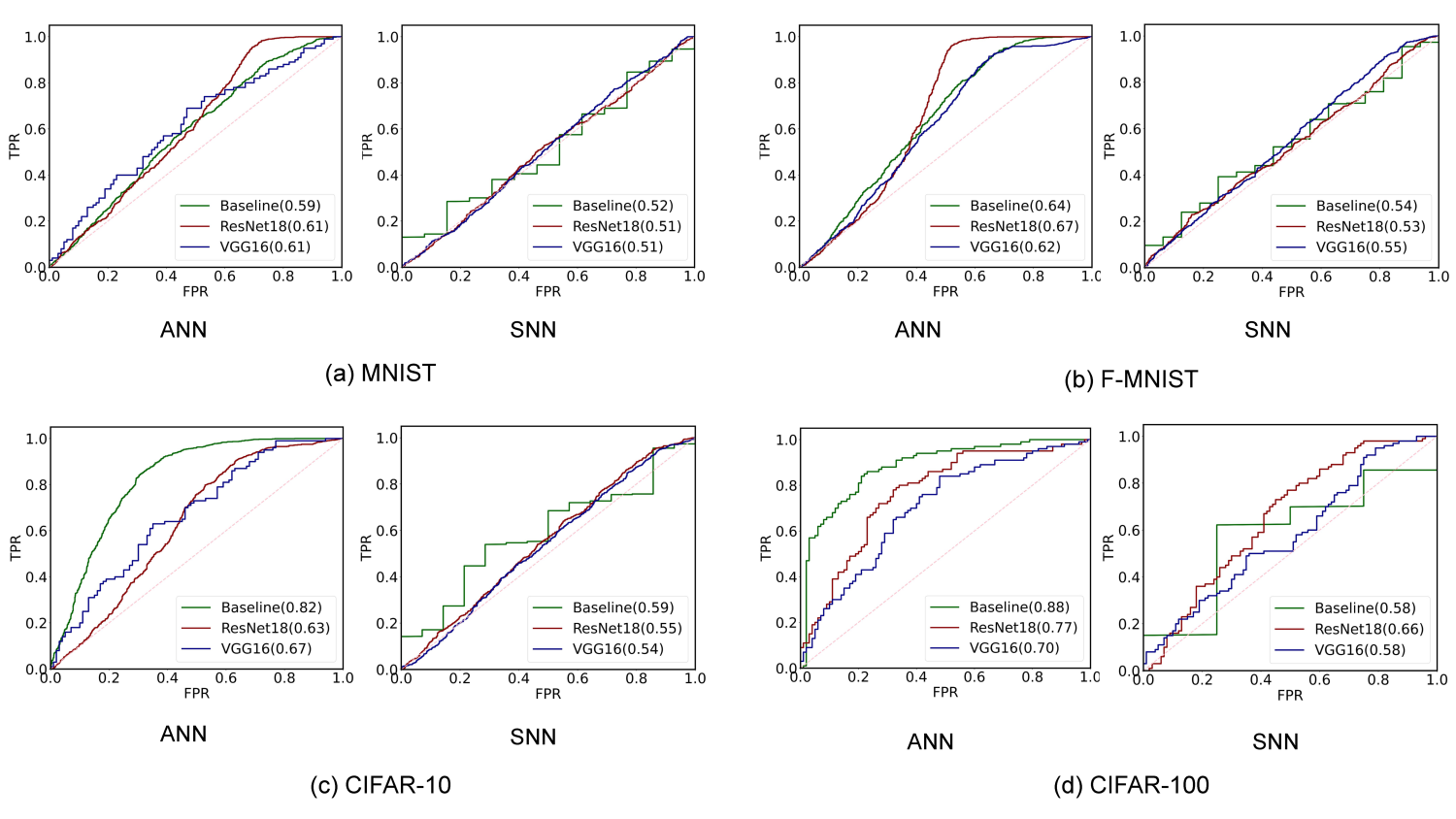}
    \caption{ROC curves comparing ANN and SNN models across different datasets: (a) MNIST, (b) F-MNIST, (c) CIFAR-10, and (d) CIFAR-100.}
    \Description{}
    \label{fig:beval}
\end{figure*}

\begin{table*}[ht!]
\centering
\caption{Comparison of ANN and SNN Resilience to MIA Across Diverse Datasets}
\label{tab:comparison_ann_snn}
\begin{tabular}{llcccccc}
\toprule
\textbf{Dataset} & \textbf{Architecture} & \multicolumn{3}{c}{\textbf{ANN}} & \multicolumn{3}{c}{\textbf{SNN}} \\
 & & \textbf{Train Acc} & \textbf{Test Acc} & \textbf{Attack AUC} & \textbf{Train Acc} & \textbf{Test Acc} & \textbf{Attack AUC} \\
\midrule
\multirow{3}{*}{\textbf{MNIST}}        & Baseline  & 99.96(±0.01)\% & 99.21(±0.03)\% & \textbf{0.59(±0.008)} & 99.95(±0.05)\% & 99.22(±0.02)\% & \textbf{0.52(±0.002)} \\
                                        & ResNet18  & 99.98(±0.10)\% & 99.57(±0.04)\% & \textbf{0.61(±0.005)} & 99.96(±0.009)\% & 99.51(±0.06)\% & \textbf{0.51(±0.006)} \\
                                        & VGG16     & 99.67(±0.05)\% & 99.50(±0.02)\% & \textbf{0.61(±0.007)} & 99.58(±0.12)\% & 99.37(±0.14)\% & \textbf{0.51(±0.004)} \\
\midrule
\multirow{3}{*}{\textbf{F-MNIST}}      & Baseline  & 99.52(±0.13)\% & 92.77(±0.20)\% & \textbf{0.64(±0.011} & 99.42(±0.32)\% & 92.44(±0.19)\% & \textbf{0.54(±0.008)} \\
                                        & ResNet18  & 98.72(±0.21)\% & 93.66(±0.15)\% & \textbf{0.67(±0.004)} & 99.12(±0.09)\% & 91.79(±0.03)\% & \textbf{0.53(±0.003)} \\
                                        & VGG16     & 97.14(±0.17)\% & 93.06(±0.20)\% & \textbf{0.62(±0.005)} & 96.53(±0.29)\% & 90.34(±0.19)\% & \textbf{0.55(±0.011)} \\
\midrule
\multirow{3}{*}{\textbf{CIFAR-10}}     & Baseline  & 99.24(±0.08)\% & 73.20(±0.43)\% & \textbf{0.82(±0.095)} & 99.13(±0.45)\% & 72.99(±0.33)\% & \textbf{0.59(±0.005)} \\
                                        & ResNet18  & 97.65(±0.51)\% & 90.81(±0.24)\% & \textbf{0.63(±0.021)} & 96.74(±0.44)\% & 86.85(±0.13)\% & \textbf{0.55(±0.011)} \\
                                        & VGG16     & 97.45(±0.41)\% & 88.74(±0.29)\% & \textbf{0.67(±0.013)} & 79.77(±0.31)\% & 74.92(±0.20)\% & \textbf{0.53(±0.003)} \\
\midrule
\multirow{3}{*}{\textbf{CIFAR-100}}    & Baseline  & 98.31(±0.19)\% & 42.46(±0.35)\% & \textbf{0.88(±0.016)} & 99.42(±0.59)\% & 39.92(±0.67)\% & \textbf{0.58(±0.019)} \\
                                        & ResNet18  & 98.66(±0.23)\% & 70.92(±0.48)\% & \textbf{0.77(±0.009)} & 88.43(±0.27)\% & 60.65(±0.74)\% & \textbf{0.66(±0.002)} \\
                                        & VGG16     & 81.61(±0.17)\% & 58.32(±0.52)\% & \textbf{0.70(±0.019)} & 60.53(±0.22)\% & 52.32(±1.36)\% & \textbf{0.58(±0.004)} \\
\midrule
\textbf{Iris}         & Baseline & 100(±0.0)\%   & 96.67(±2.34)\% & \textbf{0.77(±0.13)} & 97.04(±0.34)\% & 96.67(±0.0)\% & \textbf{0.57(±0.020)} \\
\textbf{Breast Cancer}& Baseline & 99.54(±0.04)\% & 97.37(±0.24)\% & \textbf{0.65(±0.008)} & 99.62(±0.0)\% & 96.98(±0.17)\% & \textbf{0.55(±0.017)} \\
\bottomrule
\end{tabular}
\label{{tab:comparison_ann_snn}}
\end{table*}

\begin{figure*}[ht]
    \centering
    \includegraphics[width=\linewidth, height=0.6\linewidth]{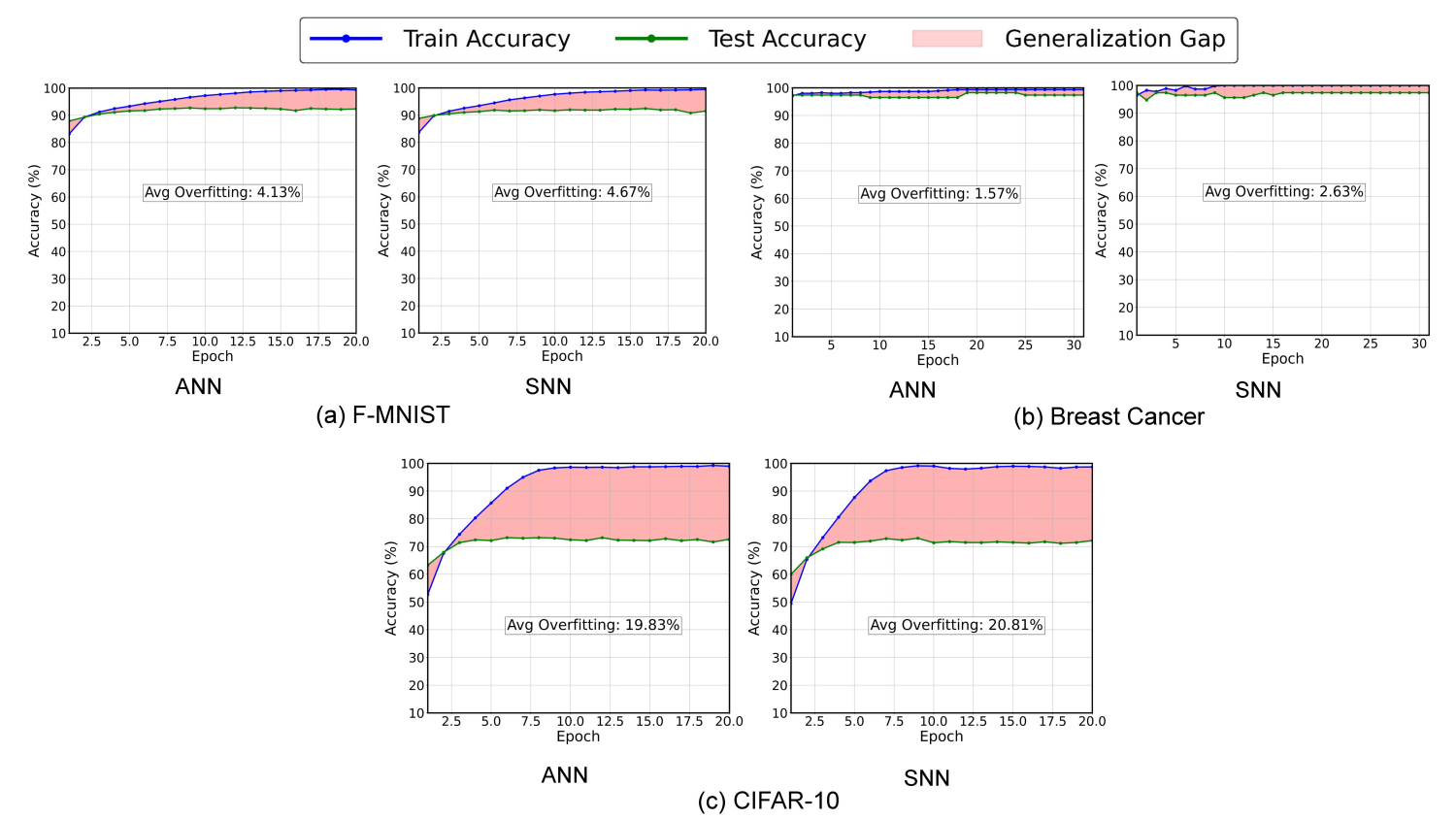}
    \caption{Overfitting analysis in ANN and SNN models across datasets: (a) F-MNIST, (b) Breast Cancer, and (c) CIFAR-10}
    \Description{}
    \label{fig:overfitting_comparison}
\end{figure*}

\begin{figure*}[ht]
    \centering
    \includegraphics[trim=0 150 0 100, clip,width=\linewidth, height=0.2\linewidth]{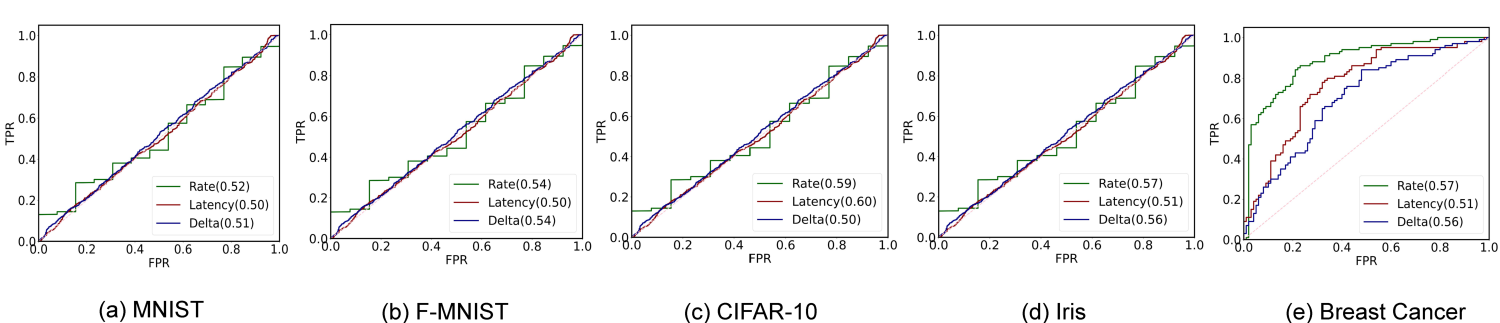}
    \caption{ROC curves showing the impact of Rate, Latency, and Delta modulation encoding under MIA for (a) MNIST, (b) F-MNIST, (c) CIFAR-10, (d) Iris, and (e) Breast Cancer datasets.}
    \Description{}
    \label{fig:ten}
\end{figure*}
\subsubsection{\textbf{Evolutionary Learning Algorithm:}}

Evolutionary algorithms, such as the Evolutionary Optimization for Neuromorphic Systems (EONS) ~\cite{schuman2020evolutionary}, are capable of enabling the rapid prototyping of SNN applications. These applications can be adapted to hardware constraints and various learning scenarios, including classification ~\cite{schuman2020automated, schuman2022opportunities} and control tasks~\cite{patton2021neuromorphic, schuman2022evolutionary}. EONS, implemented within the TENNLab framework, facilitates the co-design process between neuromorphic hardware and software, helping developers optimize neural network structures within the constraints of neuromorphic hardware. EONS interacts seamlessly with a wide variety of devices ~\cite{schuman2020evolutionary}, architectures ~\cite{10.1145/3381755.3381758}, and application ~\cite{8573122} without necessitating any modifications to its underlying algorithm. This approach, implemented within the TENNLab framework, utilizes an evolutionary algorithm to optimize neuromorphic network structure and weights. This process begins with generating a population of potential network solutions, which can be either randomly created or seeded with pre-existing networks ~\cite{schuman2020evolutionary}. The EONS process then evaluates these networks for fitness based on user-defined criteria and uses common genetic algorithm selection techniques such as tournament or fitness score to choose the most promising networks for reproduction. During reproduction, selected parent networks undergo crossover to swap segments of their structure and merging, which combines their entire structures into a single offspring. Mutation operations introduce random modifications to a network's nodes and edges, enhancing diversity and adaptation in the population. These operations are performed on a generalized network representation consisting of nodes and edges, allowing flexibility and adaptability across different hardware implementations. The entire EONS cycle, designed to optimize network parameters and structure efficiently, repeats until achieving desired performance metrics.

In our experiments we depict the impact of different EONs parameters, encoders and neuro processors inside the framework on the vulnerability of the SNN model against MIA on Iris and Breast Cancer dataset.

% \begin{itemize}
\noindent \textbf{Population Size:}
In the EONS approach, population consists of potential solutions that form the genetic pool for evolution. This initial population can be randomly generated or seeded with pre-existing solutions. Each network is evaluated, and the best-performing networks are selected for reproduction. In our experiments, we vary the population size over 50, 100, 200 and 400 to understand its impact on resilience against MIAs.

\noindent \textbf{Mutation Rate:}
The mutation rate in the EONS approach specifies the frequency at which random changes are introduced to the solutions. These mutations are essential for exploring new solution spaces and preventing the population from stagnating in local optima. Mutations can be structural, such as adding or deleting nodes and edges, or they can involve changes to parameters like thresholds or weights. In our experiments, we vary the mutation rate over 0.1, 0.5 and 0.9 to study its effects on the inherent privacy preservation in SNNs. 
 
\noindent \textbf{Crossover Rate:}
The crossover rate determines how often parts of two solutions are recombined to create new solutions, enhancing genetic diversity. In EONS, the algorithm uses a node-edge recombination method, where it mixes components from parent networks to form child networks. By varying the crossover rate, the algorithm maintains a diverse genetic pool which is essential for exploring the solution space and identifying high-performing network configurations. By varying the crossover rate over 0.1, 0.5 and 0.9, we aim to explore its impact on the resilience of SNNs against MIAs. 

\noindent \textbf{Neuroprocessors:}
We explore 3 neuroprocessors available in the TENNLab framework: 

\begin{itemize}
\item \textit{RISP} ~\cite{plank2022case}: A lightweight neuro processor employing an integrate and fire model with discrete time steps for accumulating and evaluating action potentials, suitable for networks with integer synaptic delays.
\item \textit{Caspian} ~\cite{kulkarni2021training}: Provides a high-level API and a fast spiking simulator integrated with FPGA ~\cite{kuon2008fpga} architecture, enhancing development and deployment of neuromorphic solutions in size, weight, and power-constrained environments.
\item \textit{RAVENS} ~\cite{foshie2023functional}: A versatile neuroprocessor from TENNLab with multiple implementations including software simulation, microcontroller, FPGA, ASIC~\cite{good2008asic}, and Memristive ASIC ~\cite{weiss2022hardware}(mRAVENS), catering to a wide range of computational needs in neuromorphic systems.
\end{itemize}

\noindent \textbf{Encoding Techniques:}
\begin{itemize}
\item \textit{Flip Flop Technique}: It assigns inverted percentage values in even-numbered bins, ensuring smoother transitions and preserving information about minimum values. It is effective in applications like proximity sensing in LIDAR systems ~\cite{plank2019tennlab}.
\item \textit{Triangle Technique}: It smooths input space by overlapping bins where values rise to 100\% at bin boundaries and then fall, facilitating a more gradual representation of input data. This is useful in refined control applications.
\end{itemize}
% \end{itemize}

\begin{figure*}[ht]
    \centering
    \includegraphics[width=\linewidth, height=0.6\linewidth]{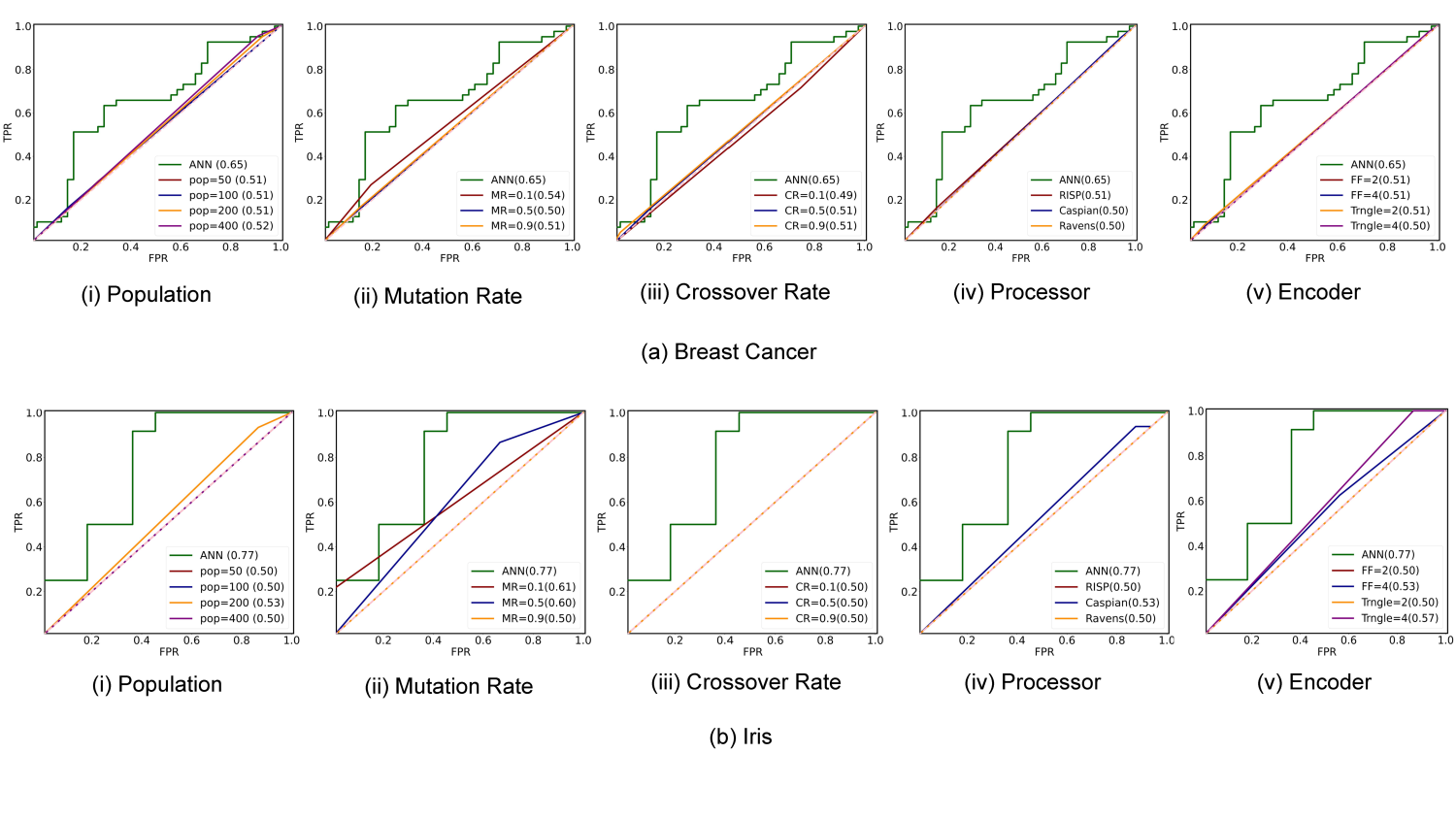}
    \caption{ROC curves comparing the impact of EONs Parameters in TennLab Framework under MIA on (a)Breast Cancer and (b)Iris Dataset}
    \Description{}
    \label{fig:tenn}
\end{figure*}
\section{Results}
\subsection{MIA Assessment}
\label{subsec:leak}
This section presents a comprehensive evaluation of ANN and SNN architectural resilience against MIAs across baseline, ResNet18, and VGG16 configurations. The privacy vulnerability is meausured using ROC-AUC metrics, where lower values indicate enhanced privacy preservation. Each experiment was repeated three times, with Table~\ref{tab:comparison_ann_snn} reporting the mean and standard deviation. The low standard deviations across all metrics indicate the statistical stability of the results.

The experimental results, illustrated in Figure~\ref{fig:beval} and Table~\ref{tab:comparison_ann_snn}, demonstrate consistent privacy advantages of SNNs while maintaining competitive accuracy. For MNIST, while both architectures achieve comparable test accuracy ($\simeq 99\%$), SNNs exhibit lower AUC values (0.51-0.52) compared to ANNs (0.59-0.61). This privacy advantage becomes more pronounced in F-MNIST, where SNNs maintain AUC values between 0.53-0.55 while achieving 90-92\% test accuracy, compared to ANNs' higher vulnerability (AUC 0.62-0.67) at similar accuracy levels.

The privacy advantage becomes more pronounced in complex datasets. For CIFAR-10, the baseline SNN maintains an AUC of 0.59 compared to ANN's 0.82, while ResNet18 and VGG16 SNN variants achieve AUCs of 0.55 and 0.53, respectively, substantially lower than their ANN counterparts (0.63 and 0.67). This trend persists in CIFAR-100, where the baseline SNN demonstrates an AUC of 0.58 versus ANN's 0.88, with ResNet18 and VGG16 SNN implementations maintaining AUCs of 0.66 and 0.58 compared to ANNs' 0.77 and 0.70.
The ROC curves in Figure~\ref{fig:beval} visually demonstrate this enhanced privacy preservation across all architectural configurations. Notably, this privacy advantage does not significantly compromise model utility, as evidenced by the competitive accuracy metrics maintained across implementations.
\noindent

\begin{tcolorbox}[
    colback=gray!10,    % Light blue background
    colframe=gray!30,  % Dark blue frame
    coltitle=black,    % Title text color
    fonttitle=\bfseries, % Bold title
    boxrule=0.8mm,     % Frame thickness
    width=\columnwidth, % Set box width to match column width
    sharp corners,     % Box style (sharp corners)
    title=Key Finding 1: % Title of the box
    ]
SNNs demonstrate enhanced privacy preservation across all architectural configurations compared to ANNs, while maintaining competitive accuracy. 
\end{tcolorbox}
\subsection{Is SNN resilience driven by ANN overfitting?}
To assess whether the improved resilience of SNNs is simply a result of ANN overfitting, the relationship between architectural privacy advantages and model generalization is examined across multiple datasets, including F-MNIST, Breast Cancer, and CIFAR-10. Overfitting is known to increase a model’s vulnerability to MIA by reducing its generalizability. Overfitted models tend to memorize training data, which increases the likelihood that an adversary can successfully infer whether a particular data point was part of the training set, thus leading to a higher MIA AUC. 

As demonstrated in Figure~\ref{fig:overfitting_comparison}, both ANN and SNN models exhibit comparable levels of overfitting, as reflected by the similar gaps between their training and testing accuracies. For example, in F-MNIST(Figure~\ref{fig:overfitting_comparison}a), the average overfitting for ANN  is 4.13\%, while for SNN, it is 4.67\%. In the Breast Cancer dataset(Figure~\ref{fig:overfitting_comparison}b), the average overfitting values are much lower, at 1.57\% for ANN and 2.63\% for SNN. These results indicate that overfitting occurs similarly in both ANNs and SNNs, ruling out overfitting in ANNs as the sole explanation for the enhanced resilience of SNNs against MIAs.

The consistency in overfitting patterns across both architectures demonstrated through comparable generalization gaps indicates that ANN's increased vulnerability to MIAs cannot be attributed to differential overfitting behavior. Despite similar levels of overfitting, SNNs consistently exhibit lower MIA AUC values compared to ANNs. This finding establishes that the enhanced privacy preservation in SNNs originates from inherent architectural characteristics rather than advantages in generalization capability, as both architectures demonstrate comparable overfitting tendencies while exhibiting distinctly different privacy vulnerabilities.

\subsection{SNN Exploration Space Assessment:}
This section examines SNN privacy characteristics across different learning algorithms, frameworks, and their associated parameters.
\subsubsection{\textbf{Surrogate Gradient Algorithm}}
In this section we analyze the effects of different encoding schemes within the snnTorch framework and the LAVA framework. 

\textbf{Impact of Encoding Schemes in snnTorch Framework:}
% \begin{figure*}[ht]
%   \centering
%   % First row - MNIST, F-MNIST, CIFAR-10
%   \begin{subfigure}{0.19\textwidth}
%     \includegraphics[width=\textwidth]{Figure/torch_enc/mnist_en_torch.png}
%     \caption{MNIST}
%     \label{fig:mnist_en_torch}
%   \end{subfigure}%
%   \hfill
%   \begin{subfigure}{0.19\textwidth}
%     \includegraphics[width=\textwidth]{Figure/torch_enc/fmnist_en_torch.png}
%     \caption{F-MNIST}
%     \label{fig:fmnist_en_torch}
%   \end{subfigure}%
%   \hfill
%   \begin{subfigure}{0.19\textwidth}
%     \includegraphics[width=\textwidth]{Figure/torch_enc/cifar10_en_torch.png}
%     \caption{CIFAR-10}
%     \label{fig:cifar10_en_torch}
%   \end{subfigure}%
% \hfill  
%   \begin{subfigure}{0.19\textwidth}
%     \includegraphics[width=\textwidth]{Figure/torch_enc/iris_en_torch.png}
%     \caption{Iris}
%     \label{fig:iris_en_torch}
%   \end{subfigure}%
% \hfill
%     \begin{subfigure}{0.19\textwidth}
%     \includegraphics[width=\textwidth]{Figure/torch_enc/bc_en_torch.png}
%     \caption{Breast Cancer}
%     \label{fig:bc_en_torch}
%   \end{subfigure}%
%   \caption{ROC curves showing the impact of Rate, Latency, and Delta modulation encoding under MIA for MNIST, F-MNIST, CIFAR-10, Iris, and Breast Cancer datasets.}
%   \label{fig:ten}
% \end{figure*}
Figure~\ref{fig:ten} illustrates the differential impact of snnTorch's three encoding schemes on SNN vulnerability to membership inference attacks. For lower complexity datasets such as MNIST (Figure~\ref{fig:ten}a) and F-MNIST (Figure~\ref{fig:ten}b), Rate and Delta encoding mechanisms demonstrate comparable privacy preservation characteristics, with AUC values converging around 0.51 and 0.53 respectively. The CIFAR-10 dataset (Figure~\ref{fig:ten}c) exhibits a broader vulnerability profile across all encoding implementations, with AUC values extending to 0.58, indicating slightly elevated susceptibility compared to simpler datasets. In the Iris dataset (Figure~\ref{fig:ten}d), the impact of encoding choice becomes more pronounced, with Delta encoding demonstrating marginally higher vulnerability (AUC = 0.56), suggesting increased encoding sensitivity in lower-dimensional datasets. Despite these variations across datasets and encoding methods, the consistently lower range of AUC values underscores the inherent privacy preserving characteristics of SNN architectures.

% Figure ~\ref{fig:ten} illustrates the influence of the three encoding schemes available in snnTorch on the vulnerability of SNNs to MIA. For the MNIST (\ref{fig:ten}a) and F-MNIST (\ref{fig:ten}b) datasets, Rate and Delta encoding exhibited similar AUC values of approximately 0.51 and 0.53, respectively, indicating a marginal influence on susceptibility to MIA. In contrast, the Iris dataset (\ref\ref{fig:ten}d) displayed a more pronounced variation in results, with the Delta encoding achieving a slightly higher AUC of 0.56. This suggests a subtle impact of encoding schemes in smaller datasets, where differences in encoding can be more impactful. For the Breast Cancer (\ref{fig:ten}d)(d), negligible differences among encoding strategies were observed, suggesting that the choice of encoder does not significantly influence MIA outcomes for these datasets. Moreover The AUC in all the experiments remains consistently low indicating high resilience towards MIA.

\textbf{Impact of LAVA Framework:}

\begin{figure}[htp]
    \centering
    \includegraphics[width=.9\linewidth]{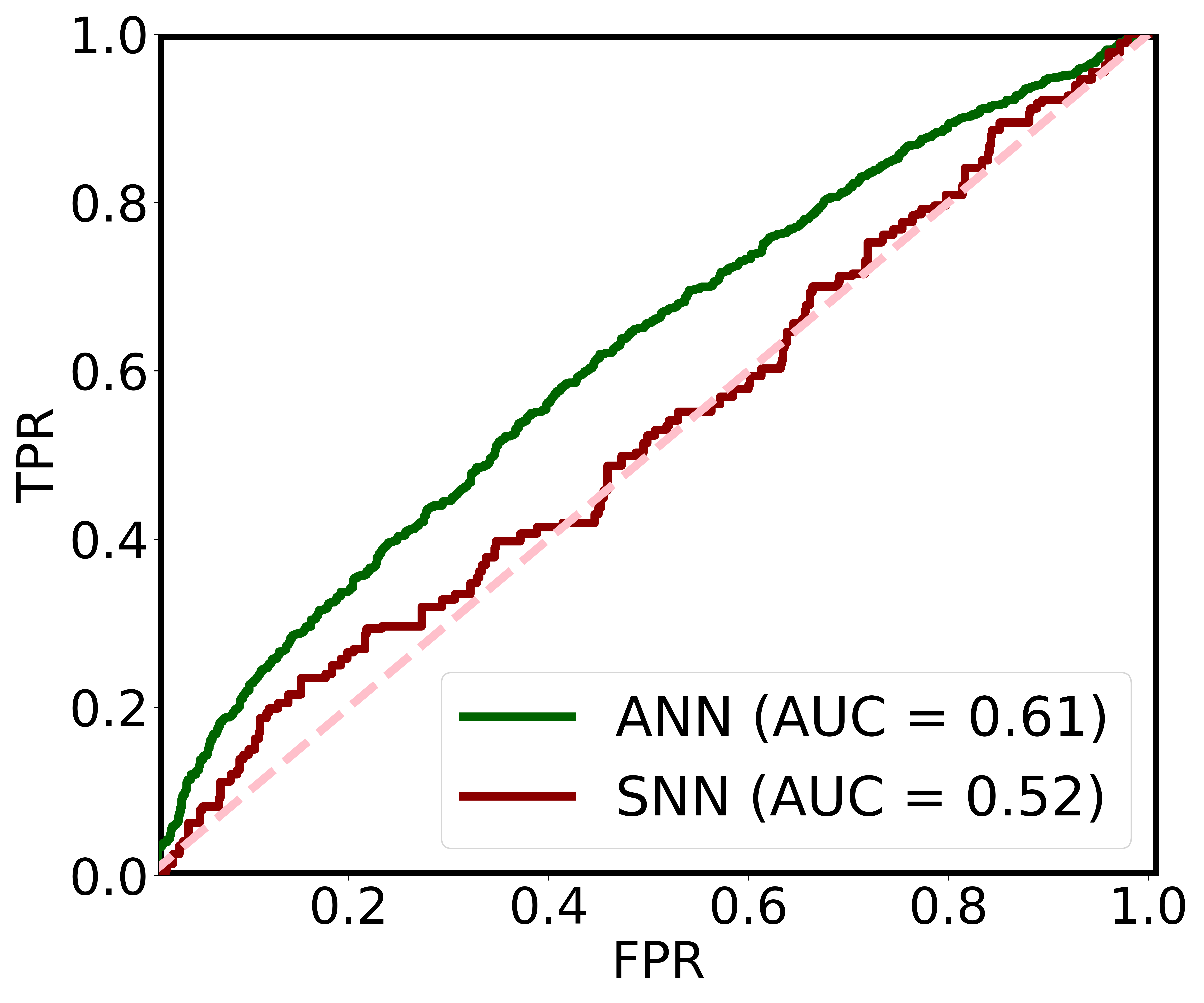}
    \caption{Impact of MIA on MNIST Dataset in LAVA Framework}
    \Description{}
    % \Description{A detailed description of the figure.}
    \label{fig:lava}
\end{figure}

The LAVA framework implementation employs a sophisticated three-layer feed-forward neural network architecture incorporating Leaky Integrate-and-Fire (LIF) neurons with pre-trained synaptic weights. The experimental results, illustrated in Figure~\ref{fig:lava}, show substantial privacy advantages in the LAVA-based SNN architecture, achieving an AUC of 0.52. This represents a significant 14.75\% reduction in vulnerability compared to the conventional PyTorch-based ANN implementation, highlighting the framework's efficacy in preserving privacy while maintaining neuromorphic computing capabilities.

\begin{figure*}[ht]
    \centering
    \includegraphics[width=\linewidth, height=0.6\linewidth]{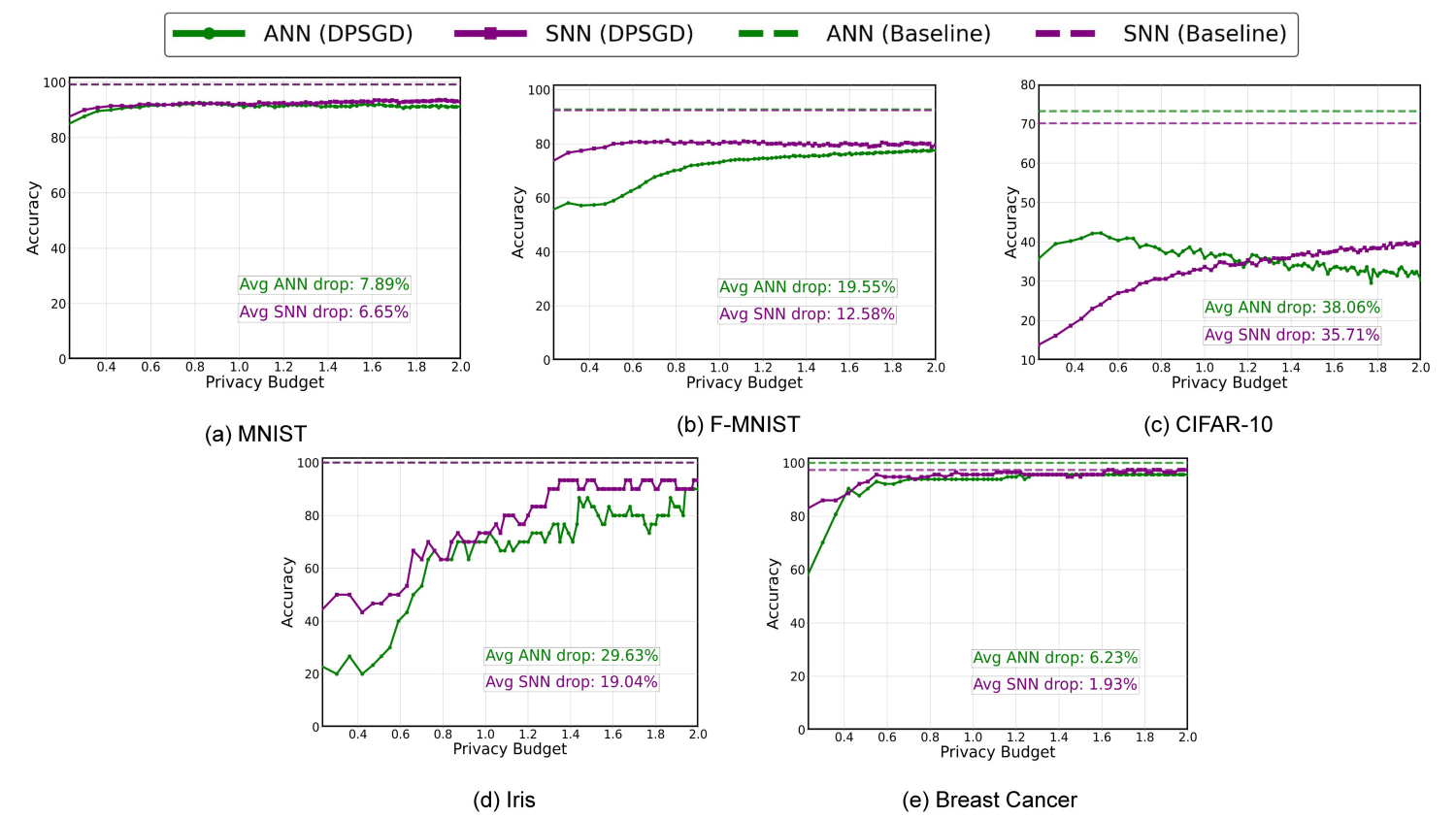}
    \caption{Model Performance over increasing Privacy Budget ($\varepsilon$) with DPSGD across (a) MNIST, (b) F-MNIST, (c) CIFAR-10, (d) Iris, and (e) Breast Cancer datasets.}
    % \description{}
    \label{fig:dp_acc}
\end{figure*}

\subsubsection{\textbf{Evolutionary Algorithm:}}
Figures~\ref{fig:tenn}a and \ref{fig:tenn}b present comprehensive evaluations of Evolutionary Optimization parameters within the TennLab Framework across the Breast Cancer and Iris datasets respectively. The Iris dataset demonstrates robust privacy preservation across multiple parametric variations: population size modifications (Figure~\ref{fig:tenn}b(i)) yield AUCs ranging from 0.50 to 0.53, while variations in crossover rates (Figure~\ref{fig:tenn}b(iii)) and mutation rates (Figure~\ref{fig:tenn}b(ii)) consistently maintain AUCs approaching 0.50, indicating stable privacy protection. The Breast Cancer dataset exhibits similarly robust characteristics, maintaining AUC values near 0.50 across diverse parameters including population size variations (Figure~\ref{fig:tenn}a(i)), crossover rate adjustments (Figure~\ref{fig:tenn}a(iii)), and different processor implementations (Figure~\ref{fig:tenn}a(iv)) such as RISP, Caspian, and Ravens.
When compared to surrogate gradient learning implementations (Table~\ref{tab:comparison_ann_snn}), where Iris and Breast Cancer datasets demonstrated AUCs of 0.57 and 0.55 respectively, evolutionary algorithms consistently achieve superior privacy preservation with significantly lower AUC values. This enhanced resilience to membership inference attacks is particularly noteworthy given the practical hardware implementation capabilities of evolutionary algorithms within the TennLab framework. The combination of robust privacy preservation and hardware practicality suggests that evolutionary optimization approaches may offer compelling advantages for secure neuromorphic computing applications.
\noindent
\begin{tcolorbox}[
    colback=gray!10,    % Light blue background
    colframe=gray!30,  % Dark blue frame
    coltitle=black,    % Title text color
    fonttitle=\bfseries, % Bold title
    boxrule=0.8mm,     % Frame thickness
    width=\columnwidth, % Set box width to match column width
    sharp corners,     % Box style (sharp corners)
    title=Key Finding 2: % Title of the box
    ]
Different SNN algorithms (surrogate gradient methods, STDP, and evolutionary optimization) demonstrate consistent privacy preservation across architectures and parameters, with evolutionary algorithms showing particular promise in combining privacy resilience with hardware practicality.
\end{tcolorbox}

\subsection{\textbf{Privacy-Utility Trade off Assessment}}
\begin{table*}[ht!]
\centering
\caption{Privacy-Utility Trade off Comparison between ANN and SNN}
\label{tab:performance_drop_comparison}
\begin{tabular}{lccc}
\toprule
\textbf{Dataset} & \textbf{Privacy Budget Range} & \textbf{ANN Avg Accuracy Drop} & \textbf{SNN Avg Accuracy Drop} \\
\midrule
MNIST & 0.22 - 2.00 & 7.89(±0.14)\% & 6.65(±0.08)\% \\
Fashion-MNIST & 0.22 - 2.00 & 19.55(±0.84)\% & 12.58(±0.78)\% \\
CIFAR-10 & 0.22 - 2.00 & 34.43(±0.20)\% & 27.87(±0.11)\% \\
Breast Cancer & 0.22 - 2.00 & 6.23(±0.33)\% & 1.93(±0.47)\% \\
Iris & 0.22 - 2.00 & 29.63(±0.81)\% & 19.04(±0.72)\% \\
\bottomrule
\end{tabular}
\end{table*}
The privacy-utility evaluation implements DPSGD across both architectures with a standardized privacy budget ($\epsilon$) range of 0.22-2. The comparative analysis, presented in Figure~\ref{fig:dp_acc} and Table~\ref{tab:performance_drop_comparison}, quantifies accuracy degradation as the differential between baseline accuracy (pre-DPSGD) and DPSGD implementation accuracy throughout training epochs. The results demonstrate enhanced utility preservation in SNN implementations across all datasets under equivalent privacy constraints.
On the MNIST dataset (Figure~\ref{fig:dp_acc}a), SNNs show an average accuracy reduction of 6.65\% compared to ANNs' 7.89\%. This pattern extends to F-MNIST (Figure~\ref{fig:dp_acc}b), where SNNs demonstrate a 12.58\% accuracy decrease versus ANNs' 19.55\%. CIFAR-10 (Figure~\ref{fig:dp_acc}c) maintains this trend with SNNs showing 27.87\% reduction compared to ANNs' 34.43\%, indicating sustained utility preservation even with increased data complexity.
The enhanced utility preservation extends to tabular datasets, with Breast Cancer (Figure~\ref{fig:dp_acc}e) showing minimal SNN accuracy degradation of 1.93\% compared to ANNs' 6.23\%. Similarly, the Iris dataset (Figure~\ref{fig:dp_acc}d) demonstrates SNN accuracy reduction of 19.04\% versus ANNs' 29.63\%. These results highlight SNNs' consistent ability to maintain utility under privacy constraints across diverse data modalities.
\noindent
\begin{tcolorbox}[
    colback=gray!10,    % Light blue background
    colframe=gray!30,  % Dark blue frame
    coltitle=black,    % Title text color
    fonttitle=\bfseries, % Bold title
    boxrule=0.8mm,     % Frame thickness
    width=\columnwidth, % Set box width to match column width
    sharp corners,     % Box style (sharp corners)
    title=Key Finding 3: % Title of the box
    ]
Under equivalent differential privacy budget, SNNs consistently demonstrate \textbf{lower accuracy degradation} than ANNs across all datasets, indicating better utility preservation while maintaining privacy guarantees. 
\end{tcolorbox}

\section{Related Work}
As machine learning systems handle increasingly sensitive data, the potential for privacy violations becomes increasingly significant. Li et al.~\cite{liu2021machine} categorize these privacy challenges into two primary areas: privacy attacks and privacy preserving techniques. Privacy attacks have emerged as a significant concern in ML due to the growing realization that models can inadvertently leak sensitive data. These attacks can broadly be classified into different types, such as model inversion attacks, model extraction attacks, MIA. Model inversion attacks~\cite{fredrikson2015model} reconstruct input data from outputs, while extraction attacks ~\cite{juuti2019prada} replicate model’s functionality without direct access to its architecture or parameters. Among these, MIAs are notable for inferring whether a specific data point was used in training. According to the survey conducted by Hu et al.~\cite{hu2022membership}, MIAs were first proposed in the context of genomic data by Homer et al.~\cite{homer2008resolving} where an attacker could identify an individual’s genome in a dataset based on summary statistics. Later, Shokri et al.\cite{shokri} introduced the first systematic MIA framework, showing how adversaries could use shadow models to infer training data membership. Salem et al. ~\cite{salem2018ml} reduced the complexity by demonstrating that a single shadow model can perform well compared to using multiple models, and they introduced metric based attacks that rely on confidence scores and entropy without the need for identical data distribution between shadow and target models. Nasr et al.~\cite{nasr2019comprehensive} further expanded MIA into white box settings, demonstrating that attackers with access to internal model parameters can perform even more effective MIAs. Melis et al.~\cite{melis2019exploiting} extended MIA to federated learning, highlighting vulnerabilities in distributed learning settings, where multiple parties collaboratively train a model. Song and Mittal~\cite{song2021systematic} highlighted the increased privacy risks in generative models such as GANs, where membership inference attacks could be carried out on synthetic data generators. Recent work by Ilyas et al. introduced LiRA (Likelihood Ratio Attack) ~\cite{carlini2022membership}, a method that further improves the accuracy of MIAs by leveraging confidence scores more effectively to distinguish training from non-training data points. In 2024, Zarifzadeh et al.~\cite{zarifzadeh2024low} introduced RMIA, a high-power membership inference attack that outperforms prior methods like LiRA and Attack-R, demonstrating superior robustness, particularly at low false positive rates (FPRs), using likelihood ratio tests.

To counteract these privacy attacks, several privacy preserving techniques have been developed.These techniques range from cryptographic approaches like homomorphic encryption(HE)~\cite{yi2014homomorphic}  and secure multi-party computation(SMPC)~\cite{zhao2019secure} to learning based defenses such as model obfuscation~\cite{9705028} and knowledge distillation~\cite{gou2021knowledge}. However, these methods often suffer from computational inefficiencies, particularly in large scale systems. Another approach is Federated Learning(FL)~\cite{li2020review}, which enables collaborative model training without sharing raw data, but remains vulnerable to attacks like MIA. Among these, Differential Privacy(DP)~\cite{dwork2006differential} has gained prominence due to its strong theoretical guarantees and practical applicability in machine learning settings. It provides a systematic framework for protecting individual data points by introducing noise during computations. In ML, this concept has been adapted through various algorithms, with DPSG)~\cite{song2013stochastic} being the most prominent which applies DP principles by adding noise to the gradient updates during training, offering a practical way to maintain privacy while training large models without significantly compromising accuracy.

While much of the data privacy research has centered around ANNs, expanding these investigations to SNNs is necessary. SNNs not only offer performance levels comparable to ANNs but also exhibit superior energy efficiency and hardware integration capabilities, positioning them as promising candidates for exploring inherent privacy features. Although privacy attacks on neuromorphic architectures remain underexplored, existing studies have yet to confirm SNNs' potential resistance to such threats. However, significant strides have been made in privacy preserving techniques within the neuromorphic domain. For instance, recent efforts by Han et al. ~\cite{han2023towards} focus on developing privacy preserving methods for SNNs, particularly utilizing FL and DP to address both computational efficiency and privacy challenges. Li et al.\cite{li2023efficient} introduced a framework that combines Fully Homomorphic Encryption (FHE) with SNNs, enabling encrypted inference while preserving SNNs' energy efficiency and computational advantages. Similarly, Nikfam et al.\cite{nikfam2023homomorphic} developed an HE framework tailored for SNNs, offering enhanced accuracy over Deep Neural Networks(DNNs) under encryption, while balancing computational efficiency. Additionally, Safronov et al.\cite{kim2022privatesnn} proposed PrivateSNN, a privacy preserving framework for SNNs that employs differential privacy to mitigate membership inference attacks, maintaining the energy efficient nature of SNNs.

\section{Conclusion}
The increasing deployment of machine learning systems in privacy sensitive domains has heightened the need for architectures that inherently protect data privacy while maintaining computational efficiency. This investigation addresses these requirements through a thorough examination of privacy characteristics in SNNs, evaluating their resilience against privacy attacks compared with traditional neural architectures. The discrete, event driven nature of spike based processing and temporal dynamics in SNNs may inherently limit information leakage compared to the continuous activations in ANNs, providing natural defense mechanisms against privacy attacks.

The experimental analysis establishes enhanced privacy preservation in SNN architectures, with attack AUC values significantly lower than traditional ANNs across all evaluated datasets (CIFAR-10: 0.59 vs 0.82; CIFAR-100: 0.58 vs 0.88). The privacy gain is particularly pronounced when employing evolutionary learning algorithms, which demonstrate superior resilience compared to gradient-based methods. Additionally, SNNs exhibit improved utility preservation under differential privacy constraints, maintaining higher accuracy levels compared to ANNs when implementing DPSGD across diverse datasets.

While these findings highlight SNNs' potential for privacy sensitive applications, particularly in resource constrained environments, they are focused on privacy preservation applications. Despite their privacy advantages, SNNs face challenges including complex training processes, potential scalability limitations, and and reliance on specialized hardware, which is necessary for optimal performance. However, within the scope of privacy preservation, their unique computational characteristics offer promising directions for secure neural architectures. Future research directions include hardware implementation analysis through Intel's Loihi neuromorphic processor, expanding privacy threat models, and integrating differential privacy mechanisms with evolutionary optimization in the TennLab framework. These investigations aim to further understand and enhance privacy preservation capabilities in neuromorphic architectures while maintaining their computational advantages.
\begin{acks}
This work was funded in part by National Science Foundation through award CCF2319619 and in part by the CHIST-ERA grant TruBrain, by the UK's Engineering and Physical Sciences Research Council (EPSRC) EP/Y03631X/1. % This work was in part funded by TruBrain Project EU-CHISTERA  (Ref-)
%This research received no specific grant from any funding agency in the public, commercial, or not-for-profit sectors.
The authors used AI-based tools, ChatGPT~\cite{chatgpt} and Claude~\cite{claude} to revise the text in all Sections to correct any typos or grammatical errors.
\end{acks}
\bibliographystyle{ACM-Reference-Format}
% \bibliography{sample-base.bib}

% \endinput
\end{document}